%% file: main.tex
\documentclass[conference]{IEEEtran}
\IEEEoverridecommandlockouts
\usepackage{cite}
\usepackage{amsmath,amssymb,amsfonts}
\usepackage{algorithmic}
\usepackage{graphicx}
\usepackage{textcomp}
\usepackage[scaled=0.84]{beramono}
\usepackage[T1]{fontenc}
\usepackage{xcolor}
\usepackage{listings}
\usepackage{hyperref}
\hypersetup{
    colorlinks=true,
    linkcolor=blue,
    citecolor=blue,      
    urlcolor=blue,
}
\usepackage{subcaption}
\definecolor{solarized@base03}{HTML}{002B36}
\definecolor{solarized@base02}{HTML}{073642}
\definecolor{solarized@base01}{HTML}{586e75}
\definecolor{solarized@base00}{HTML}{657b83}
\definecolor{solarized@base0}{HTML}{839496}
\definecolor{solarized@base1}{HTML}{93a1a1}
\definecolor{solarized@base2}{HTML}{EEE8D5}
\definecolor{solarized@base3}{HTML}{FDF6E3}
\definecolor{solarized@yellow}{HTML}{B58900}
\definecolor{solarized@orange}{HTML}{CB4B16}
\definecolor{solarized@red}{HTML}{DC322F}
\definecolor{solarized@magenta}{HTML}{D33682}
\definecolor{solarized@violet}{HTML}{6C71C4}
\definecolor{solarized@blue}{HTML}{268BD2}
\definecolor{solarized@cyan}{HTML}{2AA198}
\definecolor{solarized@green}{HTML}{859900}

\newcommand{\mcg}{\texttt{mcg}}
\newcommand{\AABB}{\texttt{AABB}}
\newcommand{\git}{\texttt{git}}
\newcommand{\class}[1]{\subsubsection{\rm \textbf{\texttt{#1}}}}

\begin{document}

\title{
    Modular Procedural Generation for Voxel Maps\\
    \thanks{%
        Research was sponsored by the Army Research Office and was accomplished
        under Grant Number W911NF-20-1-0002. The views and conclusions
        contained in this document are those of the authors and should not be
        interpreted as representing the official policies, either expressed or
        implied, of the Army Research Office or the U.S. Government. The U.S.
        Government is authorized to reproduce and distribute reprints for
        Government purposes notwithstanding any copyright notation herein.
    }
}

\author{
    \IEEEauthorblockN{1\textsuperscript{st} Adarsh Pyarelal}
    \IEEEauthorblockA{%
        \textit{School of Information} \\
        \textit{University of Arizona}\\
        Tucson, Arizona, USA \\
        \texttt{adarsh@arizona.edu}
    }
    \and
    \IEEEauthorblockN{2\textsuperscript{nd} Aditya Banerjee}
    \IEEEauthorblockA{%
        \textit{Department of Computer Science} \\
        \textit{University of Arizona}\\
        Tucson, Arizona, USA \\
        \texttt{abanerjee@email.arizona.edu}
    }
    \and
    \IEEEauthorblockN{3\textsuperscript{rd} Kobus Barnard}
    \IEEEauthorblockA{%
        \textit{Department of Computer Science} \\
        \textit{University of Arizona}\\
        Tucson, Arizona, USA \\
        \texttt{kobus@arizona.edu}
    }
}

\maketitle

\input{sections/abstract}

\begin{IEEEkeywords}
    procedural content generation, voxel, minecraft, artificial social
    intelligence, game design
\end{IEEEkeywords}

\input{sections/introduction}
\input{sections/pcg_ai_research}
\input{sections/decoupled_pcg}
\input{sections/human_spatial_cognition}
\input{sections/approach}
\input{sections/tutorial}
\input{sections/applications}
\input{sections/conclusion}
\bibliography{bibliography}

\end{document}

%% file: sections/abstract.tex
\begin{abstract}

    Task environments developed in Minecraft are becoming increasingly popular
    for artificial intelligence (AI) research. However, most of these are
    currently constructed manually, thus failing to take advantage of
    procedural content generation (PCG), a capability unique to virtual task
    environments. In this paper, we present \mcg{}, an open-source library to
    facilitate implementing PCG algorithms for voxel-based environments such as
    Minecraft.  The library is designed with human-machine teaming research in
    mind, and thus takes a `top-down' approach to generation, simultaneously
    generating low and high level machine-readable representations that are
    suitable for empirical research. These can be consumed by downstream AI
    applications that consider human spatial cognition.  The benefits of this
    approach include rapid, scalable, and efficient development of virtual
    environments, the ability to control the statistics of the environment at a
    semantic level, and the ability to generate novel environments in response
    to player actions in real time.

\end{abstract}

%% file: sections/introduction.tex
\section{Introduction}
\label{sec:introduction}

Minecraft \cite{minecraft} has recently emerged as an attractive platform for
artificial intelligence (AI) research \cite{malmo, craftassist,
DBLP:conf/ijcai/GussHTWCVS19, DBLP:journals/corr/abs-2101-11071,
DBLP:conf/cig/NguyenRGM17, tomcat, Bartlett.ea:2015, Demir.ea:2018} owing to
its popularity, ease of instrumentation and modification, and its ability to
support complex tasks in an open-world environment \cite{Szlam.ea:2019}.  A
voxel-based game environment such as Minecraft's is well-suited for designing
controlled AI experiments, as it enables researchers to access and manipulate
precise details of the environment without having to deal with complications
such as curvature or deformable objects.

However, most Minecraft environments currently used in AI research
\cite{Demir.ea:2018, Huang.ea:2020, Corral.ea:2021} are constructed manually,
thus failing to fully take advantage of a unique possibility afforded by a
virtual environment - namely, the ability to generate environments
procedurally. 

In this paper, we present \mcg{}, an library for procedurally generating voxel
environments for AI research. The library is part of the ToMCAT
project\footnote{\href{https://ml4ai.github.io/tomcat}{\tt
https://ml4ai.github.io/tomcat}} which is developing a suite of modular,
open-source AI technologies to support human-machine teaming and a
Minecraft-based testbed to evaluate them. The design of \mcg{} addresses a
number of requirements for AI research that are not sufficiently addressed by
existing approaches.

The rest of the paper is organized as follows. In \autoref{sec:pcg_advantages},
we discuss why procedural content generation (PCG) is particularly important in
the context of controlled experimental research. In \autoref{sec:decoupling},
we describe our approach in relation to other existing work on PCG for AI
research in Minecraft. In \autoref{sec:human_spatial_cognition}, we describe
how \mcg{} integrates higher-level semantics into the PCG workflow to support
human-machine teaming research. In \autoref{sec:approach}, we present the core
classes implemented in \mcg{}, and in \autoref{sec:tutorial}, a tutorial that
illustrates their usage. In \autoref{sec:applications}, we describe existing
and potential integrations with downstream AI applications. Finally, we
conclude in \autoref{sec:conclusion} by summarizing progress, noting
limitations, and describing our plans for future work.

%% file: sections/pcg_ai_research.tex
\section{PCG for AI research}
\label{sec:pcg_advantages}

There are a number of reasons to favor procedural content generation over
manual environment creation in a research context. We discuss the key ones in
this section.

\subsection{Parametric generation.}
\label{subsec:parametric_generation}

When designing an environment for human and artificial agents to perform tasks
in, it is desirable to have fine-grained control over certain features of the
environment, as they can have significant effects on task performance. For
example, in a recent simulated urban search and rescue (USAR) experiment
\cite{Huang.ea:2020}, there is a direct correlation between the size of the
number of rooms in the building and the participants' performance on the task.
Similarly, other factors that influence their performance include the number of
victims, their distribution in the building, and the presence and locations of
obstacles that inhibit navigation.

Depending on the objectives of the experiment, some of these features will be
control variables and others will be independent variables. In the case of
features that serve as control variables, it is likely that the values of these
variables are settled upon through a process of iterative experimental design.
For example, designers of a synthetic USAR task environment (e.g.,
\cite{Huang.ea:2020, Corral.ea:2021}) will likely need to try a few different
values for the number of rooms, the number of victims, the number of blockages,
etc. in order to arrive at a configuration that satisfies the experimental
requirements.

\begin{figure}
    \centering
    
    \begin{subfigure}{0.5\linewidth}
    \centering
         \includegraphics[width=.99\linewidth]{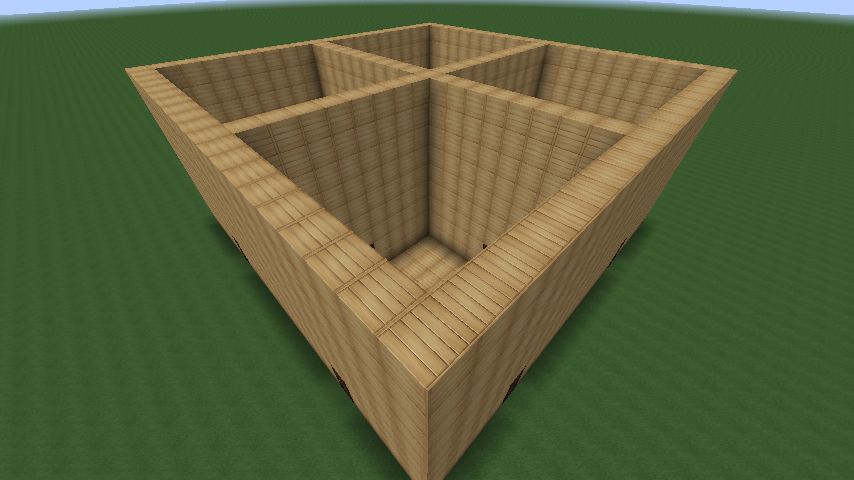}
         \caption{$2\times 2$ gridworld}
         \label{fig:2x2 gridworld}
    \end{subfigure}%
        \begin{subfigure}{0.5\linewidth}
    \centering
         \includegraphics[width=.99\linewidth]{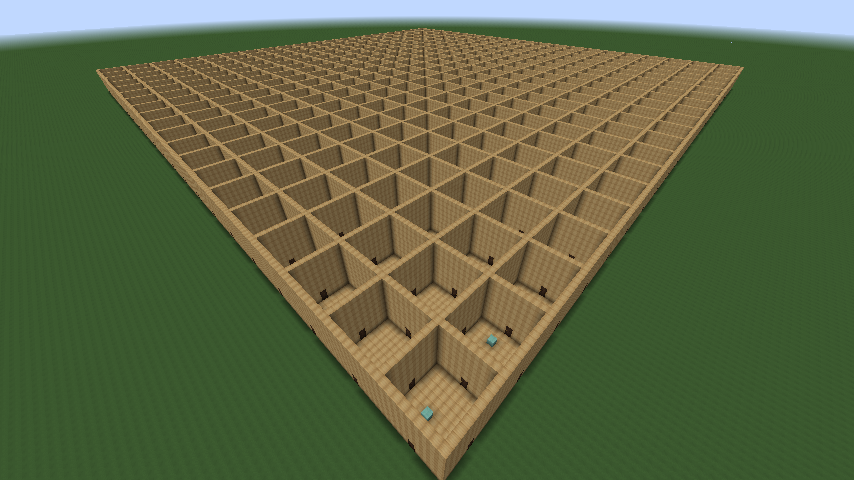}
         \caption{$20\times 20$ gridworld}
         \label{fig:20x20 gridworld}
    \end{subfigure}
    
    \caption{%
        \textbf{Gridworlds.} Two examples of gridworlds generated using \mcg{}.
        The size of the gridworld (i.e., the number of rooms) is controlled by a
        single parameter that is passed as an argument to the generator
        executable.
    }
    \label{fig:grid_world}
\end{figure}

A top-down procedural generation approach allows for rapid iteration over these
different configurations, resulting in an accelerated experimental design
process. As an example, we can go from a gridworld environment with four cells
to one with 400 cells simply by changing a single parameter (see
\autoref{fig:grid_world}). 

\begin{figure}
    \centering
    
    \begin{subfigure}{0.5\linewidth}
    \centering
         \includegraphics[width=.99\linewidth]{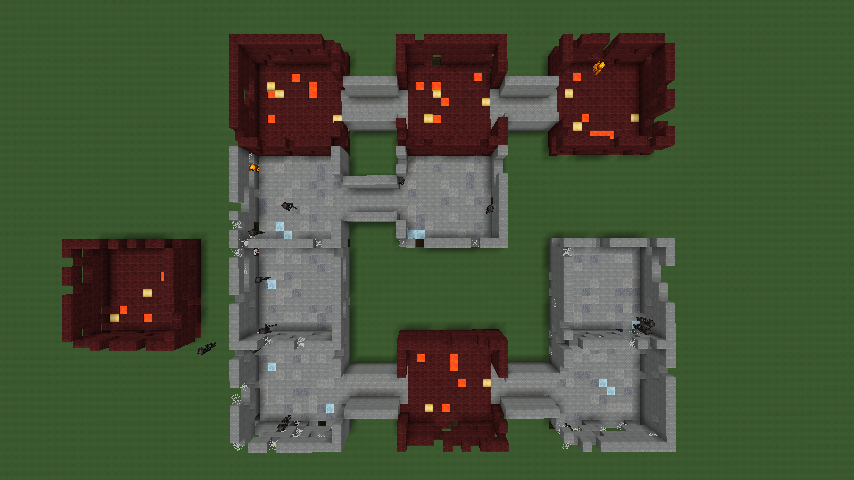}
         \caption{$4\times4$ grid, random seed = 0}
         \label{fig:dungeonworld size 16 seed 0}
    \end{subfigure}%
    \begin{subfigure}{0.5\linewidth}
    \centering
         \includegraphics[width=.99\linewidth]{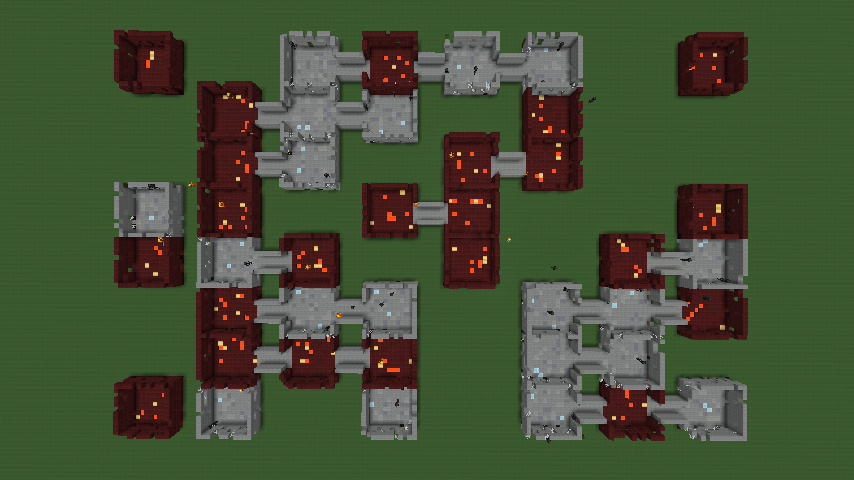}
         \caption{$8\times8$ grid, random seed = 0}
         \label{fig:dungeonworld size 64 seed 0}
    \end{subfigure}
    
    \begin{subfigure}{0.5\linewidth}
    \centering
         \includegraphics[width=.99\linewidth]{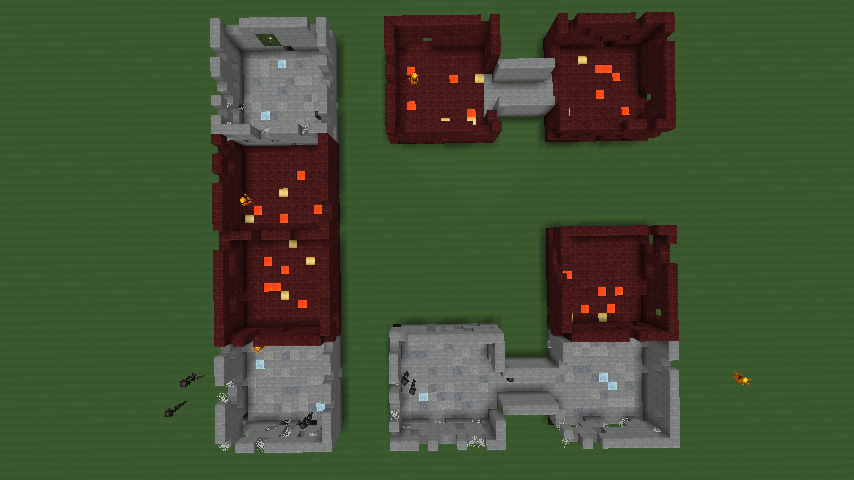}
         \caption{$4\times4$ grid, random seed = 1}
         \label{fig:dungeonworld size 16 seed 1}
    \end{subfigure}%
    \begin{subfigure}{0.5\linewidth}
        \centering
         \includegraphics[width=.99\linewidth]{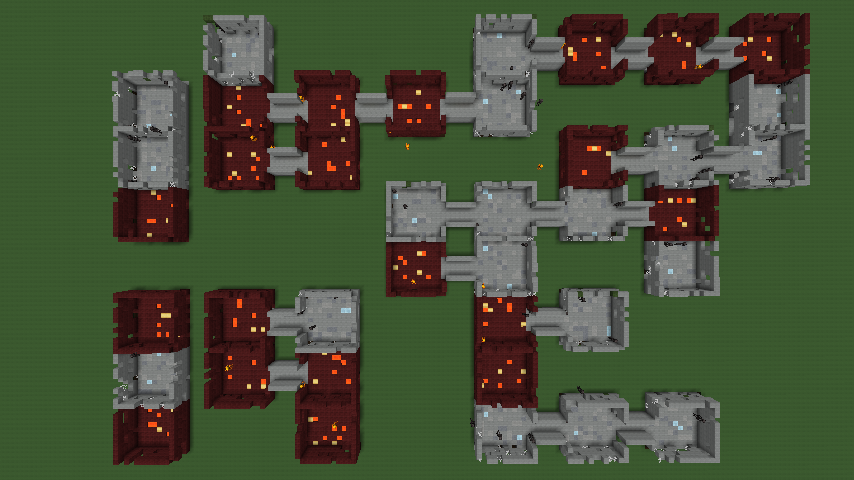}
         \caption{$8\times8$ grid, random seed = 1}
         \label{fig:dungeonworld size 64 seed 1}
    \end{subfigure}

    \caption[parindent=0.2in]{%
        \textbf{Parametric Dungeon Generation.} 
        Dungeons generated with different grid sizes ($4\times 4$ and
        $8\times 8$) and random seeds (0 and 1), using a simplified variant of
        the constraint-based algorithm described in \cite{van2013procedural}.
        We define an $N\times N$ grid, and then randomly select a number of
        grid cells to place dungeon rooms on. There are two types of rooms that
        can be `spawned' on a grid cell. The stone brick rooms (grey) contain
        diamond blocks as treasures for the player to gather, and `wither
        skeletons' as monsters for the player to battle.  The `nether' brick
        rooms (brown) contain gold block treasures and `blaze' monsters, and
        are more difficult to navigate, as they have randomly placed patches of
        lava that the player can fall into if they are not careful. The stone
        brick rooms also contain randomly placed spiderwebs on the walls for
        aesthetic purposes.  Finally, we connect the rooms with stone brick
        corridors.

        Notably, we take a modular approach to generation - each type of of
        room is defined using a class that inherits from the \AABB{} class
        described in Sec. \ref{subsubsec:aabb} - the treasures and monsters are
        placed when the room is instantiated, rather than after all the rooms
        have been placed and interconnected. This gives us a lot of
        flexibility. For example, the class that corresponds to the stone brick
        rooms could be instantiated for some of the cells in the gridworld
        environment in \autoref{fig:grid_world} (thus setting up an interesting
        surprise for players who may not have been expecting to battle
        skeletons in a USAR scenario!).
    }
    \label{fig:dungeon_world}
\end{figure}

Similarly, randomizing the environment is important for certain types of
experiments. Indeed, this can be viewed as a special case of parametric
generation, with the random seed serving as the parameter to be varied. In
\autoref{fig:dungeon_world}, we show four possible dungeon layouts obtained by
varying the size and random seed parameters passed to a dungeon world
generator implemented using \mcg{}. 

\subsection{Controlling the statistics of generated scenes} 
\label{subsec:statistics}

In experiments where environments matter, they should not be limited to what
designers intuitively create. For example, simply saying ``maybe this complex
needs a cul-de-sac'' to investigate how that would affect participants'
frustration levels in USAR scenarios does not lead to careful science. The
feature needs to be appear in different contexts with well defined
probabilities to guard against confounding effects between the presence of the
feature and its possibly unique or statistically biased context. While this
could be arranged manually, doing so is tedious and harder to get right. On the
other hand, specifying the environments procedurally forces researchers to be
more clear about the logic of their experiment, makes development easier as
modifications can be studied rapidly, and better supports extensions, scaling,
and repeatability. 

\subsection{Changing the world in response to player action} 
\label{subsec:reactive}

The ToMCAT project is mandated to support a large range of behavioral
psychology studies into team performance. It is easy to construct scenarios
where we will need to change the Minecraft environment as a function of what
the human players are doing, their affective state, brain activity (using EEG
or fNIRS equipment), and where they are in the virtual environment. As such, it
is not predictable where, and hence precisely what the needed modification will
be. For example, suppose we are interested in how robust a team's performance
is to unexpected events such as a room collapse specifically when the team
members cannot see each other. To do this, we would need to programmatically
collapse a room in the task environment according to a defined stochastic
process, but only when they are out of view of each other. In particular, the
room that needs to collapse is not known in advance.  While we could have a
single room collapse model manually specified for each room, the awkwardness of
this approach amplifies with each additional changeable feature.  For example,
a second room collapse entails being ready for any two rooms to change in any
order.

Currently, \mcg{} supports this kind of programmatic change in a scalable
manner, since the high-level representation it outputs provides semantic
`hooks' that a Minecraft mod can leverage to preserve the desired collapse
models across task environments with different numbers of rooms, varied room
layouts, and scenarios in which different numbers of rooms will need to be
collapsed. While the existing executables built using \mcg{} write files to
disk, it is conceivable that the library could be used to provide a `PCG as a
service' program that can generate complex environment specifications on the
fly.

\subsection{Task environments as code}
\label{subsec:scm}

Initially, the task environments for the ToMCAT project were built manually,
rather than procedurally. As the experimental design evolved, it became
quickly apparent that we would need to keep track of the different
versions of the task environment, so that (i) it would be easy to revert back
to an older version if necessary, and (ii) the version could be added to the
provenance for a given experimental trial.

While source code management (SCM) systems (e.g., \git{} \cite{git}) excel at
version management, they are designed for text files rather than binary files
such as the ones corresponding to the manually-constructed Minecraft
environments.  Managing binary files (especially ones that are liable to change
frequently, as is the case with an iterative experimental design process) with
SCM systems results in the source code repository getting bloated, hindering
developer productivity and collaboration.

For these reasons above, we decided to maintain versions of the binary save
files on a public server that developers could access. However, we soon
realized that this approach had its own share of problems, the most troublesome
of which was the fact that every change to the environment required manually
uploading the updated binary files to our server, making it easy for the
environment to get out of sync with the source code of the ToMCAT Minecraft
mod.

Switching to procedurally generated environments addresses all of these issues
- taking an `environment as code' approach allows us to effectively leverage
SCM systems to manage environment versions, making tasks like reverting back to
a previous version or comparing versions much easier than with manually
constructed environments. In addition, we no longer have to worry about
manually synchronizing the task environment and the code that interacts with it
- the SCM system takes care of this for us.

%% file: sections/decoupled_pcg.tex
\section{Decoupled PCG for rapid iteration}
\label{sec:decoupling}

There exists prior work on procedural generation to support AI experiments in
Minecraft. Notably, Project Malmo \cite{malmo} provides a declarative XML-based
API to specify `missions', including the parametric generation and placement of
individual blocks, entities, and simple structures such as spheres and cuboids.
In addition, it supports implementing custom procedural generation algorithms
as Java classes in the Malmo mod and exposing them via the XML API.

However, the procedural generation capabilities in Project Malmo are tightly
coupled with Minecraft itself - to view the results of a procedurally
generation algorithm implemented in a Malmo class, one would need to recompile
the Malmo mod and relaunch Minecraft with it loaded. This process is fairly
slow and does not fit well into a workflow that involves the need for rapid
iteration through environments. Ideally, there should be a way to visualize the
outputs of procedural generation without having to launch the full game.

In contrast, our library is \emph{decoupled} from Minecraft - it outputs
declarative specifications of an environment that can be consumed by other
downstream applications, including a Minecraft mod that can translate the
specification into in-game entities, blocks, and structures. The data
structures and algorithms in our library are general enough that they could be
used for any other voxel-based environment - the only things that would need to
be modified are the labels for the block and entity types.

%% file: sections/human_spatial_cognition.tex
\section{Connection to human spatial cognition}
\label{sec:human_spatial_cognition}

One of the primary motivations for developing \mcg{} was to explicitly inject
high-level semantic information (labels of relevant locations and structures,
layout and topology, etc.) into the generation process. The Malmo API and the
Minecraft Forge event bus give us access to low-level information about the
positions of the player and individual blocks in the environment. However,
humans tend to reason about their environment at a high level of abstraction.
For example, a human carrying out tasks in a Minecraft environment will tend to
think about `rooms' and `buildings' rather than the individual blocks that the
structures are comprised of. 

Furthermore, humans rely on high-level spatial representations of their
environment for navigation \cite{Gallistel:1990}. These representations have
been hypothesized to take the form of Euclidean maps \cite{OKeefe.ea:2016} or
graph-like representations \cite{Warren:2019}, though there is evidence that
both representations may be simultaneously maintained and used in different
contexts\cite{Peer.ea:2021}. 

One way to incorporate high-level semantics is by manually specifying location
boundaries, labels, and hierarchies after an environment is built.  However,
this method is prone to human error and bottlenecked by the time it takes for
humans to annotate regions and identify area boundaries in a pre-built map.
This method will certainly not scale to large or stochastically generated maps.

In contrast, the \mcg{} library takes a `top-down' approach, producing the
following machine-readable representations \emph{simultaneously}, in lockstep
with each other.

\begin{itemize}

    \item \textbf{High-Level Representation (HLR)}: Also known as a `semantic
        map', this is a JSON file that contains information about the labels
        and locations of areas, their connections with each other, and their
        hierarchical relationships, as well as the \texttt{Entity} and
        \texttt{Object} instances in the environment. This representation
        supports linking explicitly to human spatial cognition.

    \item \textbf{Low-Level Representation (LLR)}: This is a JSON file that
        contains low-level information about all the blocks and entities in the
        environment. Among other uses, this representation can be consumed by a
        Minecraft mod to generate an environment procedurally, but with the
        actual PCG algorithms offloaded to the \mcg{} library.
        
\end{itemize}

These specifications can then be used by other programs, some examples of which
we describe in greater detail in \autoref{sec:applications}.

%% file: sections/approach.tex
\section{Approach}
\label{sec:approach}

The \mcg{} library provides a set of core components that can be extended and
composed to design rich voxel-based task environments. In this section, we
briefly describe these components\footnote{The full documentation for the C++
    API can be found at
\href{https://ml4ai.github.io/tomcat/cpp_api/index.html}{\tt
https://ml4ai.github.io/tomcat/cpp\_api/index.html}.} and the design philosophy
behind them, followed by a tutorial on how to use the library.

\subsection{Core classes}

\class{Pos}
This class represents a point in the three-dimensional integer
lattice $\left(\mathbb{Z}^3\right)$ \cite{Maehara.ea:2018} - or in other words,
a vector in a 3D Euclidean space with the constraint that its Cartesian
components are restricted to integer values, reflecting the voxelated nature of
the Minecraft environment. 

\class{AABB}
\label{subsubsec:aabb}

\begin{figure}
    \centering
    
    \begin{subfigure}{\linewidth}
        \centering
         \includegraphics[width=\linewidth]{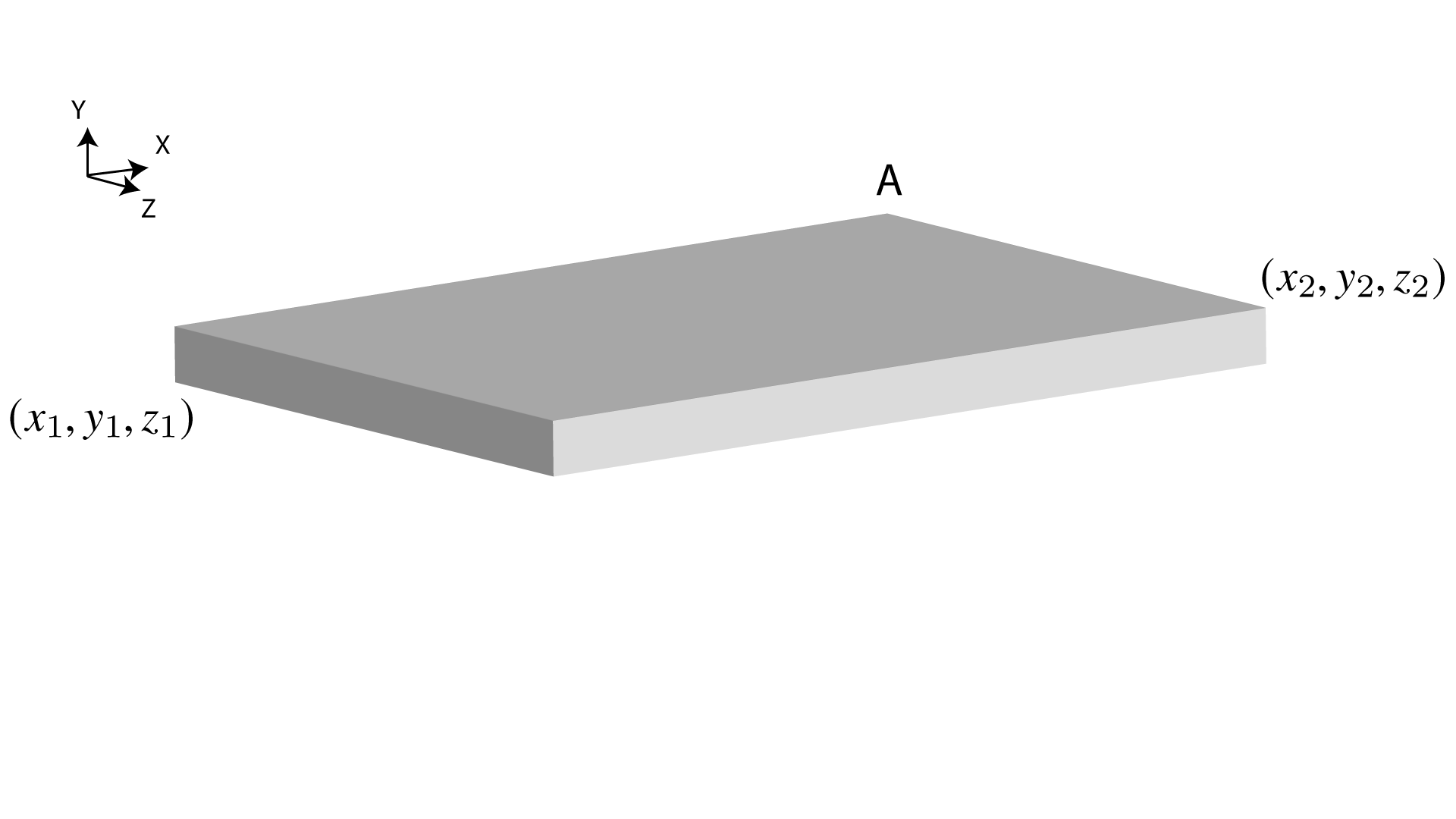}
         \caption{%
             An AABB represented by the pair of coordinates $(x_1, y_1, z_1)$
             and $(x_2, y_2, z_2)$ which represent the top left and bottom
             right corners of the cuboid respectively.
         }
         \label{fig:aabb}
    \end{subfigure}
        
    \begin{subfigure}{\linewidth}
        \centering
        \includegraphics[width=\linewidth]{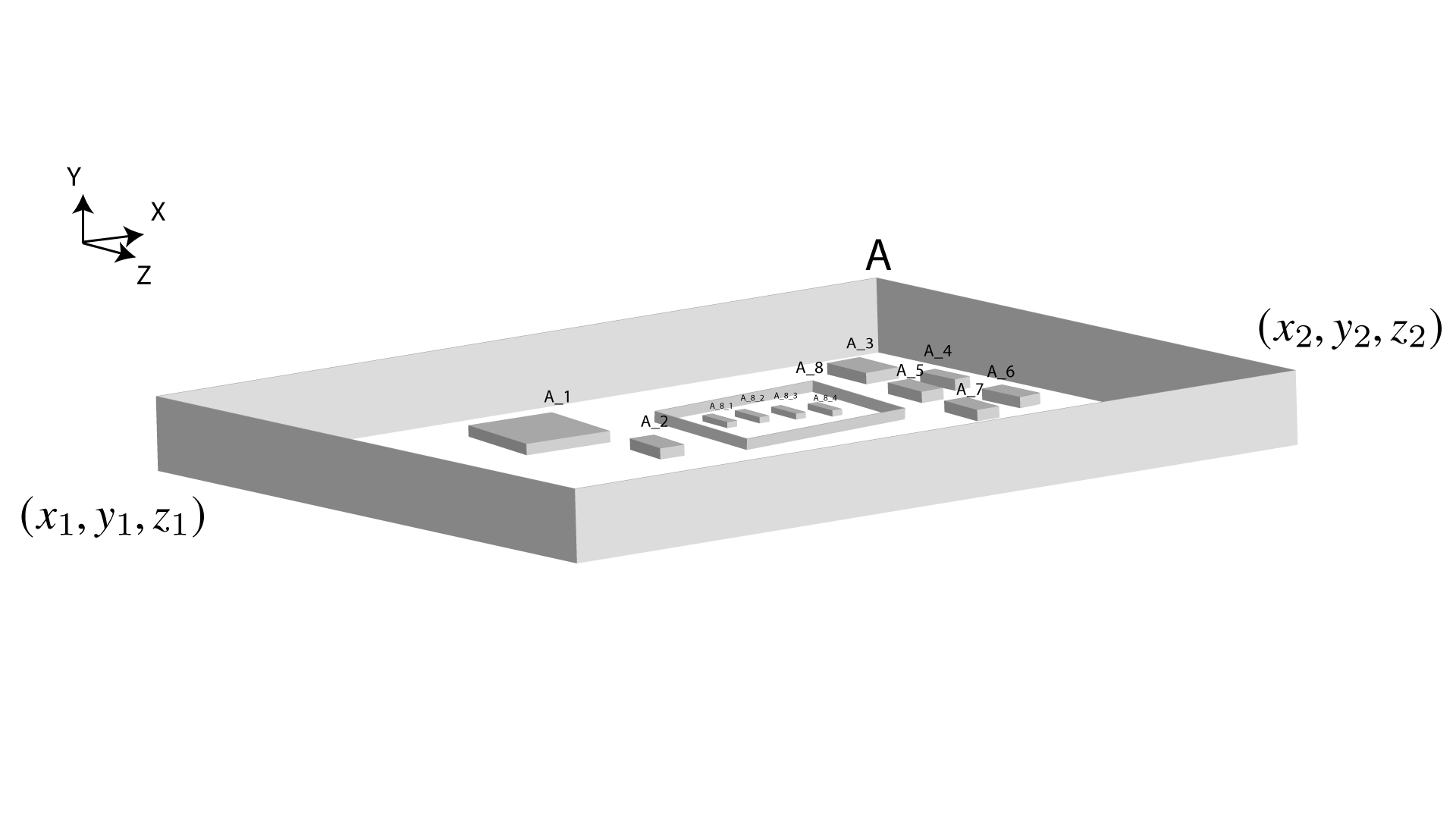}
        \caption{%
            An AABB can itself can contain one or more AABBs inside it, as
            shown in this example - the AABB in \autoref{fig:aabb} contains a
            number of child AABBs. Grouping AABBs into a hierarchy gives us the
            benefit of encapsulation. For example, if a set of AABBs is contained
            within a parent AABB and each child AABB is defined relative to the
            parent's boundary or to the other child AABBs, one need only move
            the parent AABB to automatically move the child AABBs into the
            correct relative positions as well.
        }
        \label{fig:group}
    \end{subfigure}

    \begin{subfigure}{\linewidth}
        \centering
         \includegraphics[width=\linewidth]{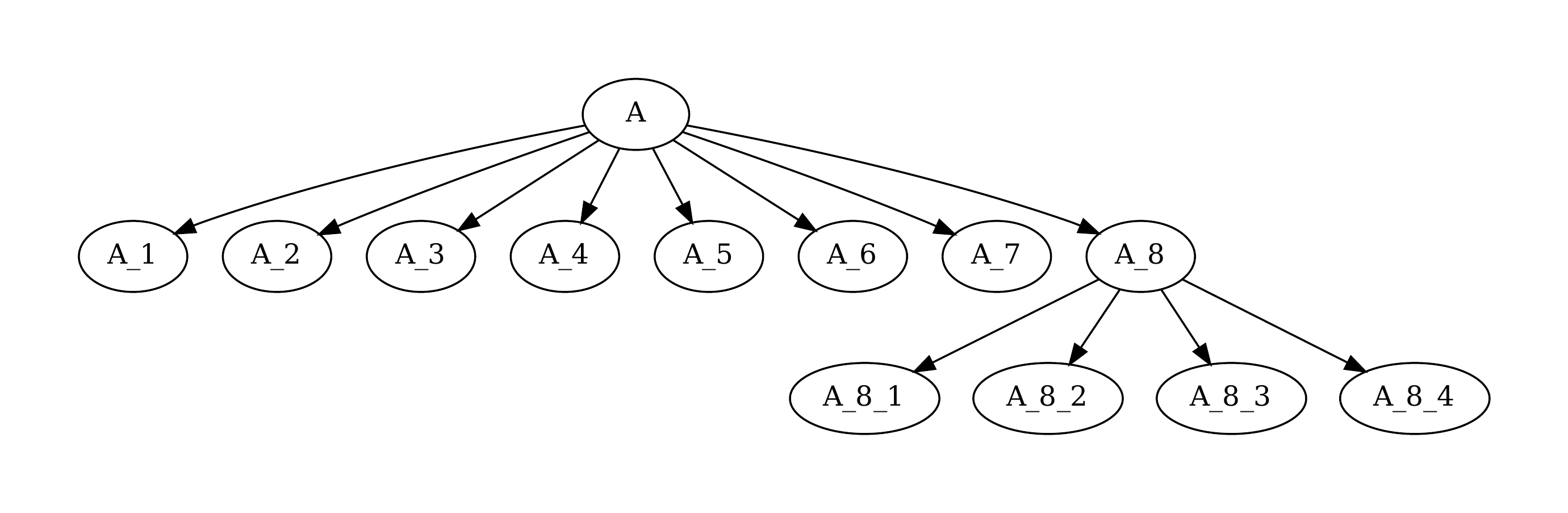}
         \caption{%
             A graph-like representation is automatically created
             and kept track of by \mcg{}, to facilitate linking with human
             spatial cognition (\autoref{sec:human_spatial_cognition}). The
             graph above shows the parent-child relationships between the AABBs
             in \autoref{fig:group}.
         }
         \label{fig:group_graph}
    \end{subfigure}
    
    \caption{%
        \textbf{Axis Aligned Bounding Boxes (AABBs).} We use AABBs as the
        semantic building blocks of \mcg{}.
        They can be combined and nested into larger AABBs to form complex
        structures that are addressable as semantically meaningful
        components. In this figure, we show an example of a location hierarchy
        formed by nesting AABBs.
    }
    \label{fig:design_philosophy}
\end{figure}

An axis-aligned bounding box (AABB) is an elementary cuboidal structure that
can be efficiently represented using a pair of 3D coordinates that correspond
to vertices on opposite ends of one of its space diagonals. The \AABB{} class
instantiates an empty cuboidal space that effectively serves as a blank canvas
to implement PCG algorithms in. For example, one could implement Perlin noise
generation \cite{DBLP:conf/siggraph/Perlin85, santamaria2014procedural} to add
water blocks within an \AABB{} made of grass to create a water body, or
constrained growth algorithms \cite{green2019organic, lopes2010constrained}.
AABBs of different types can be defined using \mcg{} by subclassing the \AABB{}
class, and can be manipulated, nested, and combined to produce complex
structures (see \autoref{fig:design_philosophy}).

\class{Block}

This class represents a single Minecraft block with a given material and
position. It allows for fine-grained placement of individual blocks in the game
environment - for example, the diamond and gold treasure blocks in
\autoref{fig:dungeon_world}.

\class{Entity}

This class represents an entity that is to be placed in the task environment,
The constructor for this class takes two required arguments - the type and
position (a \texttt{Pos} object\footnote{While Minecraft allows for the
    components of the Cartesian coordinates of \texttt{Entity} objects to be
double-precision floating point numbers, we restrict them to be integers in
\mcg{} for simplicity.}) of the entity, and a set of optional arguments
corresponding to the different equipment types that a Minecraft entity can
have. We use this class to generate and place the wither skeletons and blazes
in the dungeons in \autoref{fig:dungeon_world}.

\class{World}

This class represents the overall environment, and contains the lists of the
\AABB{}s, \texttt{Block}s, \texttt{Entity}s, \texttt{Object}s and
\texttt{Connection}s in it as class attributes.

\class{Object}

An object represents a \texttt{Block} with some additional semantics.  An
instance of this class contains information about an id, type, and
\texttt{Block} associated with the \texttt{Object}. It is particularly useful
in cases where a Block holds some semantic meaning, like the victim blocks in
\cite{Huang.ea:2020}.

\class{Connection}

The \texttt{Connection} class represents a spatial connection between \AABB{}s.
It is meant to encompass a variety of structures that can fall under the
semantic label of `connection' - it can be used to represent both `point-like'
connections (e.g., a door between rooms) and `extended' connections (e.g., a
corridor that connects two locations). 
Some care must be taken with \texttt{Connection} objects. Unlike entities and
objects in an \AABB{} that will be automatically moved to the appropriate
locations when an \AABB{} undergoes spatial translation, \texttt{Connection}
objects will not be so reliably updated. However, since \texttt{Connection}
objects can be stored in \AABB{} objects and \texttt{World} objects, a
potential workaround for this issue would be to instantiate connections
dynamically when \AABB{} are moved around.


%% file: sections/tutorial.tex
\section{Tutorial}
\label{sec:tutorial}

In this tutorial, we showcase some of \mcg{}'s capabilities and demonstrate how
to use the classes described in \autoref{sec:approach}.  The goal of this
tutorial will be to create a house with two rooms and a zombie in a purely
programmatic manner. 

\subsection{World Setup}
We start by creating an empty world. In a file named
\texttt{mcg\_tutorial.cpp}, we will add the following:

\begin{lstlisting}
#include "mcg/World.h"
using namespace std;

class TutorialWorld : public World {
  public:
  TutorialWorld() {};
  ~TutorialWorld(){};
};

// Create the world and write the JSON output to file.
int main(int argc, char* argv[]) {
  TutorialWorld world;
  world.writeToFile(
      "semantic_map.json",
      "low_level_map.json");
  return 0;
}
\end{lstlisting}

This minimal program will produce two files, \texttt{semantic\_map.json} and
\texttt{low\_level\_map.json} that correspond to the HLR and LLR respectively.
The LLR is used by the \texttt{WorldBuilder} class in the ToMCAT Minecraft mod
to construct the environment. At this point, the generated world will be
empty\footnote{To actually view the generated environment in Minecraft, you can
use the script \texttt{tools/run\_mcg\_tutorial} that is included in the
repository.}.

\subsection{Creating a room}

We will now add single room to this world. To do so, define a class called
\texttt{Room} that extends \AABB{}, and whose constructor takes a string
identifier and a \texttt{Pos} object that represents the top left corner of the
room \footnote{By the `top left' corner of an AABB, we mean the corner with the
lowest values of the \emph{X}, \emph{Y} and \emph{Z} coordinates}.

\begin{lstlisting}
...
#include "mcg/AABB.h"

class Room : public AABB {
  public:
  Room(string id, Pos& topLeft) : AABB(id) {}
  ~Room(){};
};
...
\end{lstlisting}

Note that we invoke the superclass constructor, which sets all instances of
\texttt{Room} to have a material type of \texttt{blank} and the coordinates of
the top left and bottom right corner set to $(0,0,0)$. To turn this blank
\texttt{Room} into an actual room, modify the constructor as shown below.

\begin{lstlisting}
  ...
  Room(string id, Pos& topLeft) : AABB(id) {
    // Set the base material to be 'log'
    this->setMaterial("log");

    // Define the object's boundaries
    Pos bottomRight(topLeft);
    bottomRight.shift(5, 4, 5);
    this->setTopLeft(topLeft);
    this->setBottomRight(bottomRight);
  }
  ...
\end{lstlisting}

This will give us a room made of logs with a $6\times6$ block base and a height
of 5 blocks. Finally, modify the \texttt{TutorialWorld} constructor to place a
\texttt{Room} instance at $(1,3,1)$\footnote{We set the \emph{y}-coordinate to
    3 to match the level of the ground in the Minecraft world.}. The
    constructor should now look as follows:

\begin{lstlisting}
  ...
  TutorialWorld() {
    Pos topLeft(1, 3, 1);
    auto room1 = make_unique<Room>("room_1", topLeft);
    this->addAABB(move(room1));
  };
  ...
\end{lstlisting}

\subsection{Adding details}
\label{subsec:adding_details}

\begin{figure}
  \centering
  \includegraphics[width=.99\linewidth]{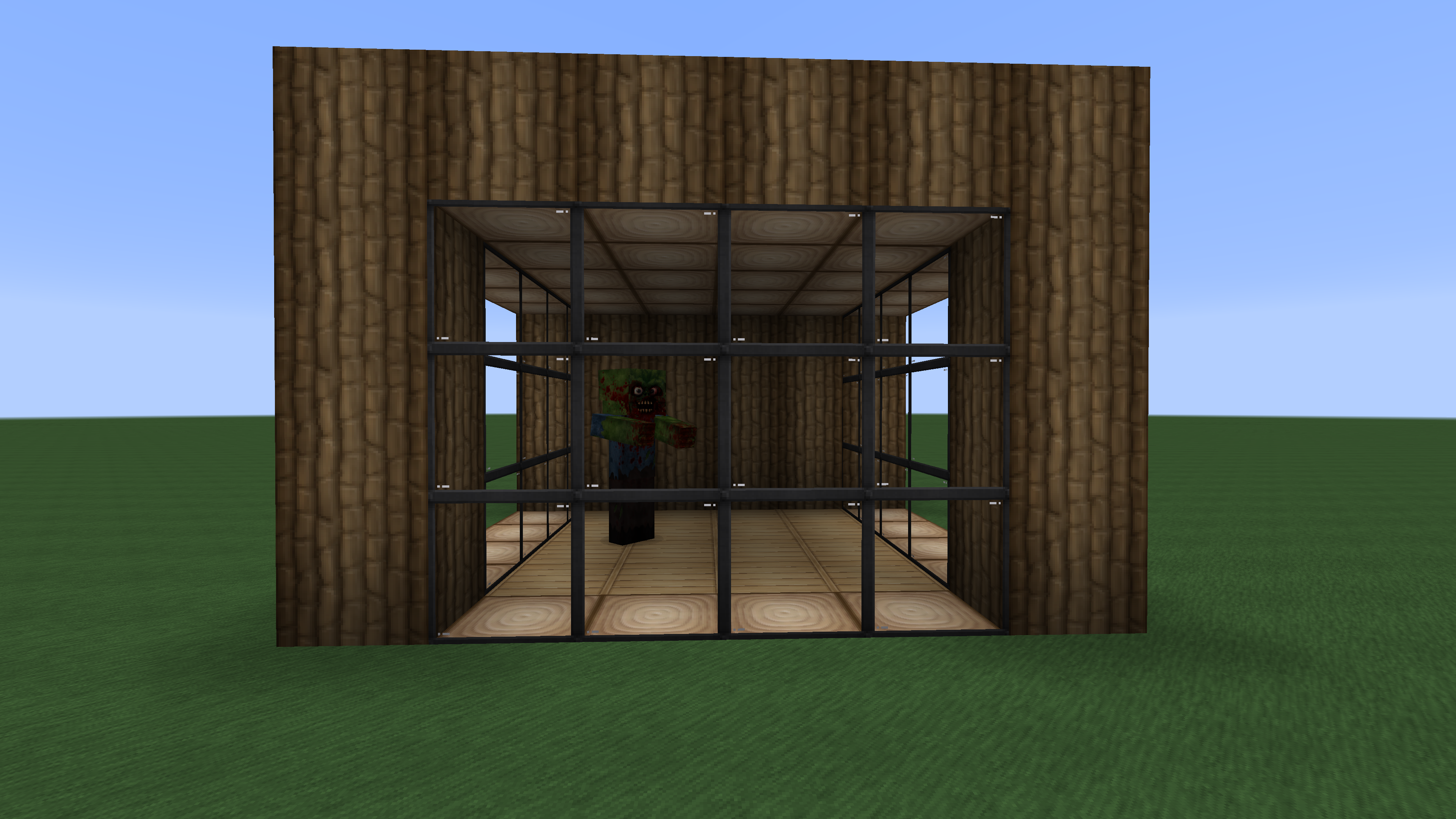}
  \caption{%
    This figure shows the \texttt{TutorialWorld} \texttt{Room} with all the
    details added in \autoref{subsec:adding_details}. It now has a floor made
    of planks, windows on three sides, a roof, and a zombie.
  }
  \label{fig:detailed_tut_house}
\end{figure}

We will now add some details to this room - namely, a floor, roof, windows, and
a zombie (see \autoref{fig:detailed_tut_house}). This is accomplished by adding
the following to the \texttt{Room} constructor.

\begin{lstlisting}
    ...
    // The floor should be made of planks
    this->generateBox("planks", 1, 1, 0, 4, 1, 1);

    // Add windows made of glass
    this->generateBox("glass", 0, 5, 1, 1, 1, 1);
    this->generateBox("glass", 5, 0, 1, 1, 1, 1);
    this->generateBox("glass", 1, 1, 1, 1, 0, 5);

    // Add a roof (will be made of logs)
    this->hasRoof = true;

    // Add a zombie
    mt19937_64 gen; // Random number generator engine
    Pos randomPos = this->getRandomPos(gen, 1, 1, 1, 2, 1, 1);
    auto zombie = make_unique<Entity>("zombie", randomPos);
    this->addEntity(move(zombie));
    ...
\end{lstlisting}

Note that the coordinates passed to the \texttt{generateBox} method are
relative to the \AABB{} itself, so we do not need to respecify them when
placing a second room.

\subsection{Multiple rooms}
\label{subsec:multiple_rooms}

Finally, we will combine two \texttt{Room} instances to create a house. To do
this, we create a second \texttt{Room} instance and add both rooms to an
enclosing \AABB{}. This is done by adding the following code to the
\texttt{TutorialWorld} constructor:

\begin{lstlisting}
    ...
    Pos topLeft(1, 3, 1);
    auto room1 = make_unique<Room>("room_1", topLeft);
    auto room2 = make_unique<Room>("room_2", topLeft);
    room2->shiftX(5);

    auto house = make_unique<AABB>("house");
    house->addAABB(move(room1));
    house->addAABB(move(room2));
    this->addAABB(move(house));
    ...
\end{lstlisting}

\begin{figure}[h]
    \centering
    \includegraphics[width=.99\linewidth]{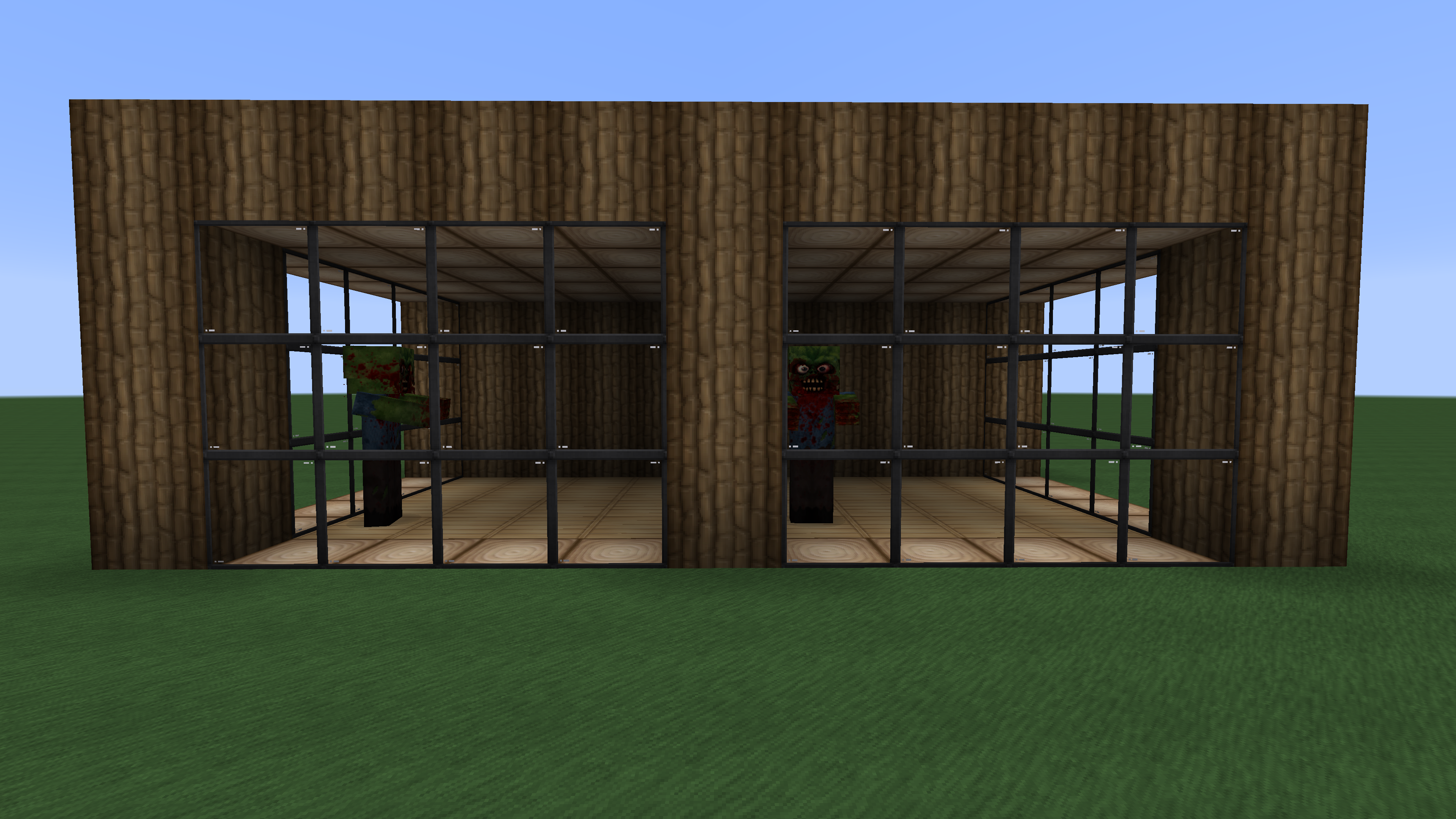}
    \caption{%
        This figure shows the completed \texttt{TutorialWorld} house. It has
        two rooms, each with the details added in
        \autoref{subsec:multiple_rooms}.
    }
    \label{fig:complete_tut_house}
\end{figure}

\autoref{fig:complete_tut_house} shows the completed \texttt{TutorialWorld}
with a house that has two rooms. The code for this tutorial along with
instructions on how to compile and run it can be found in the
\texttt{libs/mcg/examples/mcg\_tutorial} folder\footnote{All folder names are
    given relative to the root of the version of the ToMCAT repository
    corresponding to the tag \texttt{2021\_IEEE\_CoG}
    (\href{https://github.com/ml4ai/tomcat/releases/tag/2021_IEEE_CoG}{\tt
https://github.com/ml4ai/tomcat/releases/tag/2021\_IEEE\_CoG}). The linked
version of the tutorial code contains more detailed comments than that shown
here for additonal clarity.}.

%% file: sections/applications.tex
\section{Applications}
\label{sec:applications}

The machine-readable high-level and low-level representations that are
simultaneously produced by \mcg{} can be consumed by a number of downstream
applications. We divide them into two broad categories: \emph{agents} and
\emph{non-agents}. 

As mentioned earlier, \mcg{} is being developed as part of the ToMCAT project
\cite{tomcat}, which is in turn part of DARPA's Artificial Social Intelligence
for Successful Teams (ASIST) program \cite{Elliot:2019}.  The testbed developed
for the program \cite{Corral.ea:2021} publishes real-time measurements of each
participant's state, environment, and actions in Minecraft to an MQTT message
bus \cite{mqtt}, along with data from other sources such as physiological
sensors and pre and post-task questionnaires.

The term `agent' is a fairly overloaded one. However, in this paper, we use the
term to refer to programs that the various ASIST performer teams are developing
that subscribe to topics on the message bus, process it in a streaming manner,
and publish their outputs back to the message bus, where they can potentially
be used by other, downstream agents.

\begin{figure}
    \centering
    \includegraphics[width=\linewidth]{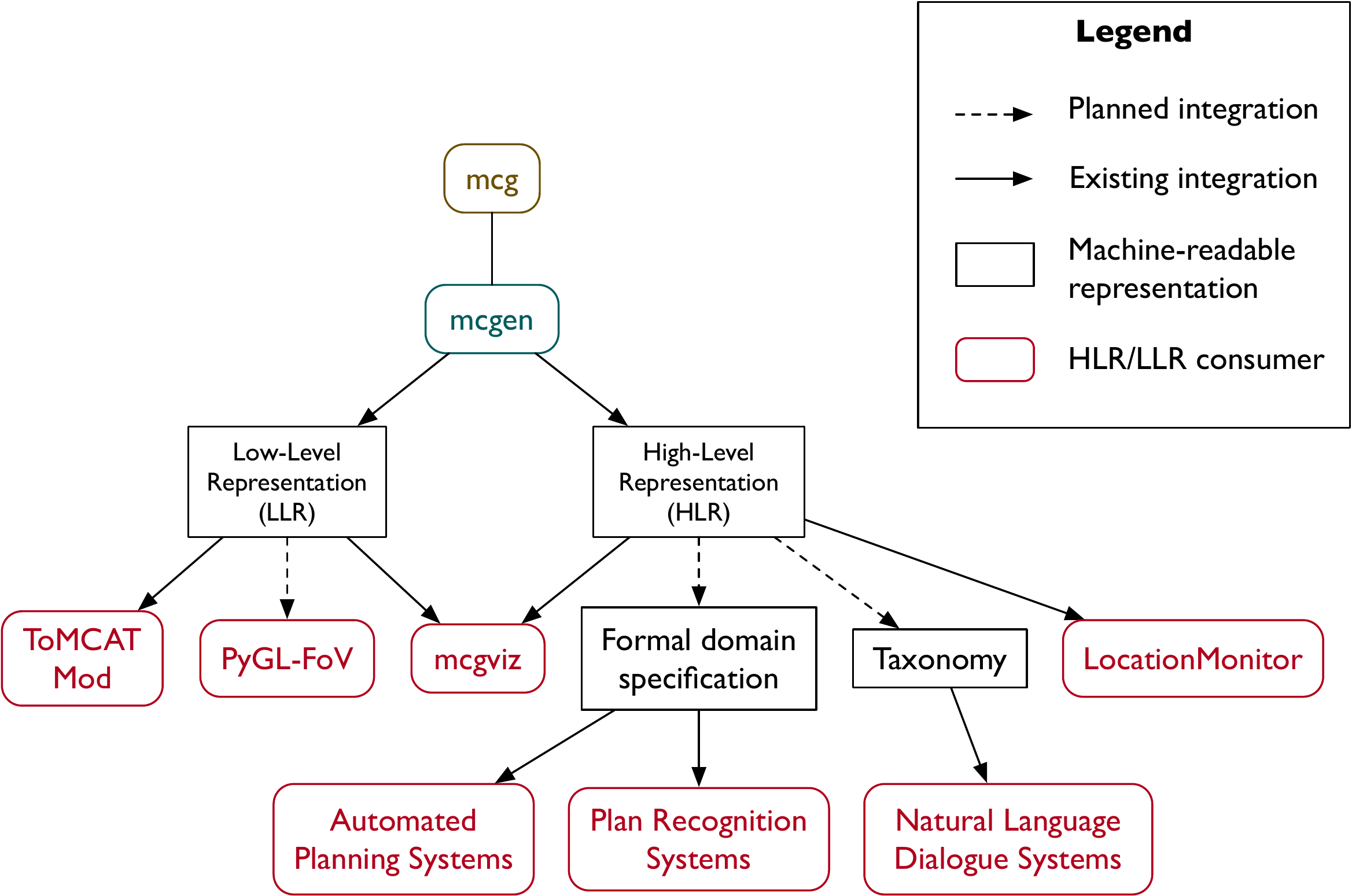}
    \caption{%
        \textbf{Applications}. In this figure, we show the existing and
        potential integrations of \mcg{}. The agents (the PyGL-FoV
        and location monitor agents, natural language dialog systems, plan
        recognition and planning systems) can potentially exchange data with
        each other, but we do not show those connections in order to focus on
        how the HLR and LLR are used by the agents. 
    }
    \label{fig:applications}
\end{figure}

In contrast, we use the term `non-agents' to mean software components that are
not primarily designed to process streams of information. The potential
applications and their relation to the high and low level representations are
shown in \autoref{fig:applications}.

\subsection{Non-agents}

\subsubsection{\texttt{mcgen}}

The \texttt{mcgen} program is being developed for ToMCAT experiments, and is
representative of generator executables that can be built using \mcg{}. The
environments shown in \autoref{fig:grid_world}, \autoref{fig:dungeon_world},
and \autoref{fig:mcgviz} are all generated using \texttt{mcgen}.

\newcommand{\mcgviz}{{\rm\texttt{mcgviz}}}

\subsubsection{\mcgviz{}}

\begin{figure*}
    \centering
    
    \subcaptionbox{Graph representation\label{fig:zombieworld_graph}}{
         \includegraphics[width=0.33\linewidth]{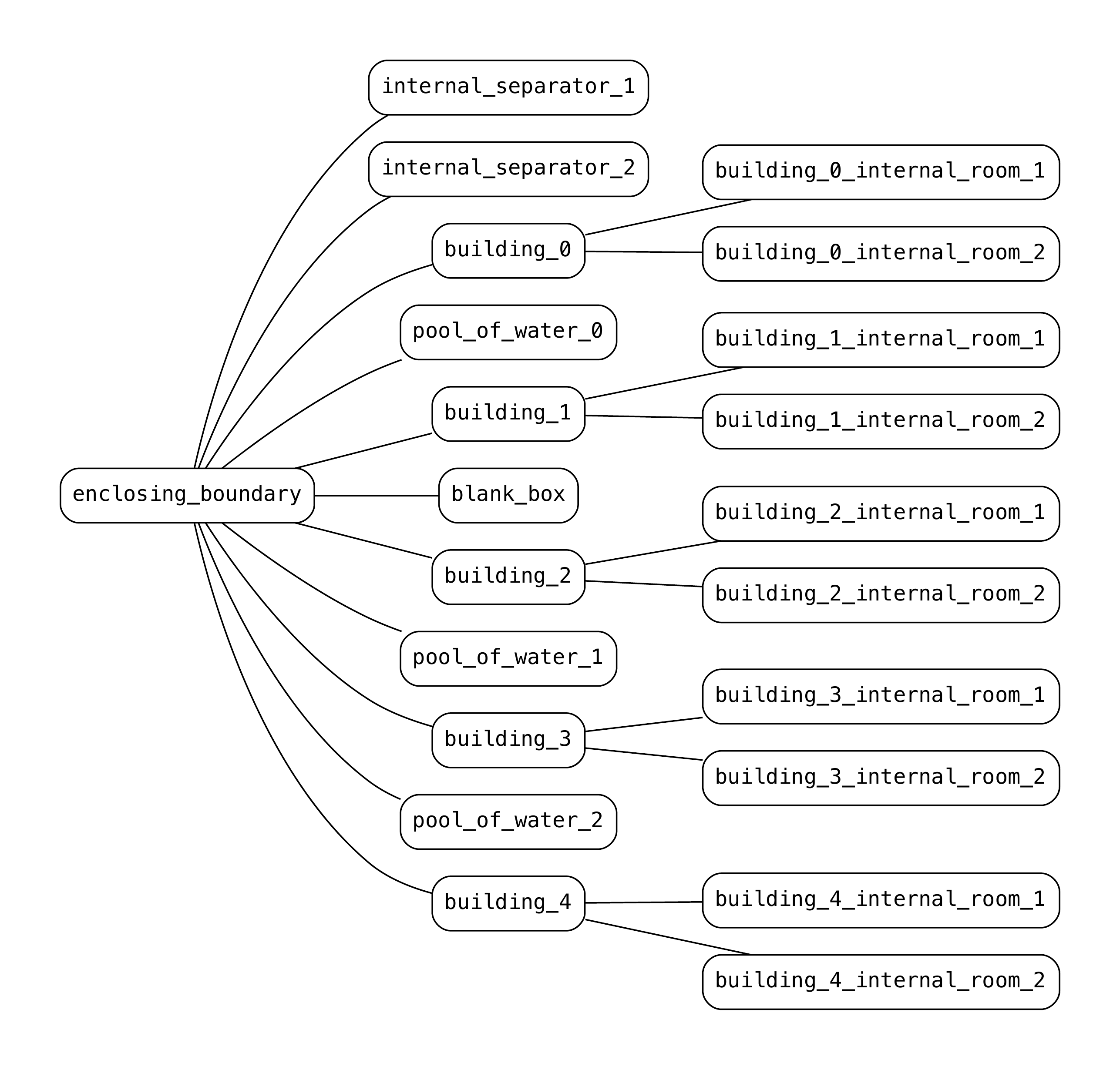}
    }
    \subcaptionbox{Blueprint\label{fig:zombieworld_bp}}{
         \includegraphics[width=0.3\linewidth]{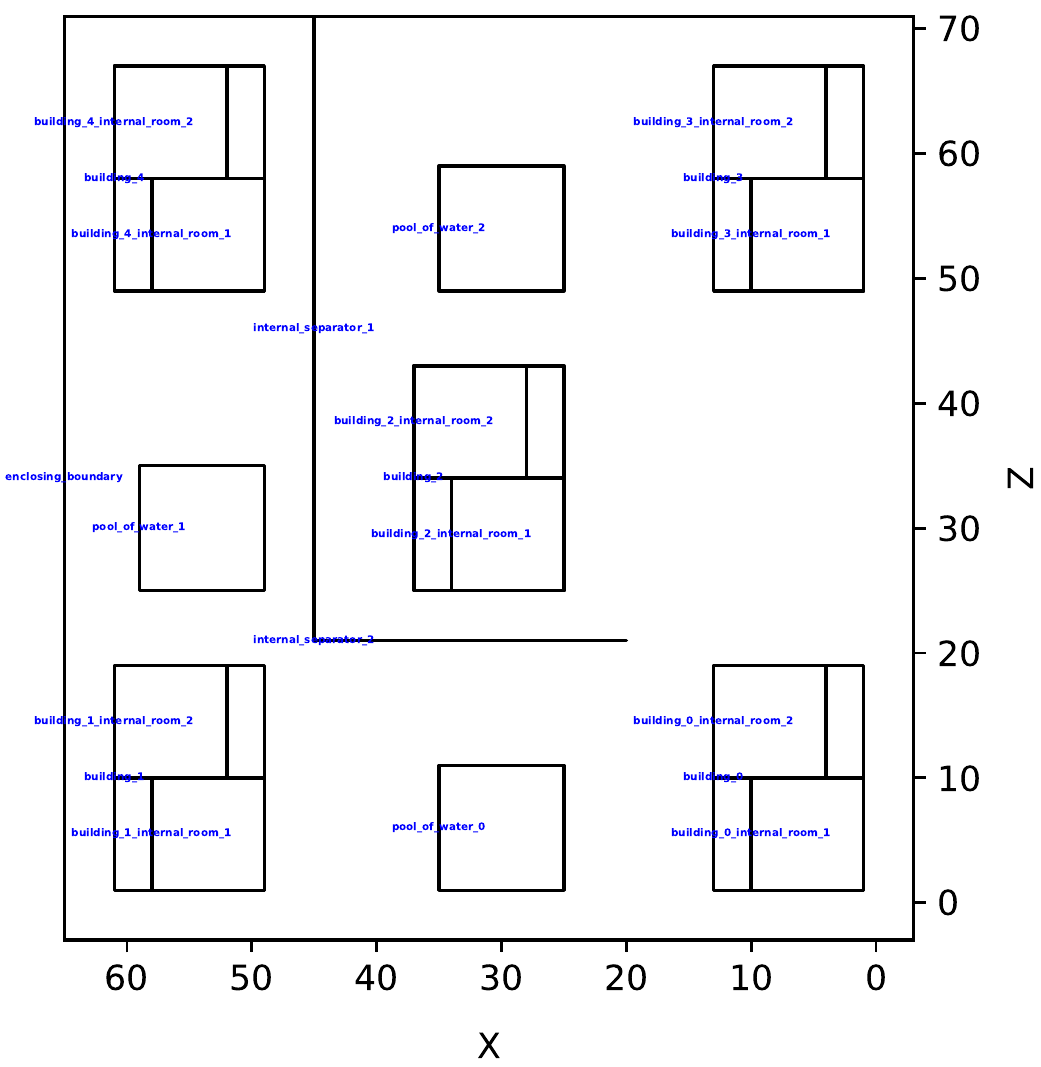}
    }
    \subcaptionbox{Screenshot\label{fig:zombieworld_screenshot}}{
         \includegraphics[width=0.29\linewidth]{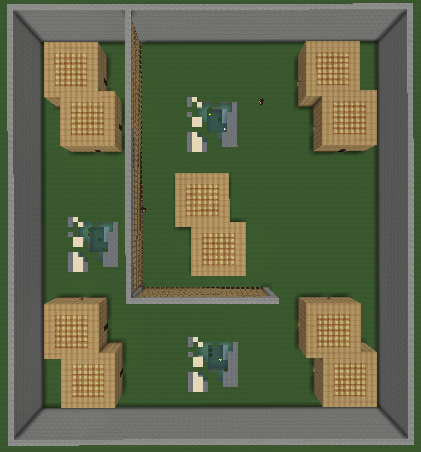}
    }
    
    \caption{%
        Three different views of the same \textbf{ZombieWorld} environment.
        To construct this environment, we define an enclosing boundary \AABB{} to which we
        add two main components, a \texttt{ZombieWorldGroup} and a
        \texttt{ZombieWorldPit}. The \texttt{ZombieWorldGroup} is a building
        with two rooms, internal doors and a zombie and villager, while the
        \texttt{ZombieWorldPit} is a pool of lava or water, depending on how it
        is initialized.  We place these structures in a $3\times3$
        grid such that when it is time to place a \texttt{ZombieWorldPit}, we
        randomly choose whether to fill it lava or water or skip placing
        the pit entirely. We also add two internal walls within the enclosing
        boundary.
    }
    \label{fig:mcgviz}
\end{figure*}

\mcgviz{} is a Python script included with \mcg{}, that takes the HLR and
LLR output by \mcg{} to produce visualizations of the environment. Using the
HLR, it can construct either a graph structure (e.g.
\autoref{fig:zombieworld_graph}) or a `blueprint' style visualization showing a
top-down view of the AABBs in the generated environment (see
\autoref{fig:zombieworld_bp}). It can also combine the LLR and HLR to provide a
more detailed map with individual colored patches corresponding to the
different types of blocks in the environment.

\subsubsection{ToMCAT mod}

The ToMCAT mod is a Minecraft mod that builds upon the Malmo mod \cite{malmo}
with additional functionality for human-machine teaming research. Currently,
the LLR produced by \mcg{} is consumed by the ToMCAT mod to construct the
in-game environment (see \autoref{fig:zombieworld_screenshot}).

\subsection{Agents}

\subsubsection{LocationMonitor}

The LocationMonitor agent developed by IHMC \cite{ihmc} for the ASIST program
uses a `semantic map' - that is, the HLR output by \mcg{} - to construct an
internal representation of named locations (e.g. rooms, hallways) with their
boundaries and connections to other named locations. Using this internal
representation, it monitors the player's position (in Cartesian coordinates)
and publishes a message to the message bus whenever a player goes from one
named location to another.

\subsubsection{PyGL-FoV}

PyGL-FoV \cite{pygl_fov} is an agent that uses observations of the Cartesian
coordinates of the player, the pitch and yaw of the gaze vector of their
Minecraft avatar, and a low-level representation of the environment to compute
whether certain blocks of interest are visible on the player's screen at any
given time.

\subsubsection{Dialog Systems}

There is growing interest in using Minecraft as an environment to develop
dialog-enabled artficial agents \cite{craftassist,
DBLP:conf/acl/JayannavarNH20, tomcat}. Dialog systems such as the ToMCAT
DialogAgent\footnote{\url{https://github.com/clulab/tomcat-text}} rely on a taxonomy
of concepts to ground natural language extractions to. In order to ground to
specific locations that are referred to by participants (especially if they
have been provided a blueprint with the location labels beforehand), the
taxonomy will need to incorporate location names - the HLR produced by
\mcg{} can be used to automate the construction of the spatial portion of
the taxonomy.

\subsubsection{Planning and Plan Recognition Systems}

The HLR produced by \mcg{} contains information about named locations and their
connections to each other - it can be used to automatically construct portions
of formal planning domain and problem specifications related to spatial
information.  For example, the connectivity information in the HLR can be used
to automatically generate a number of predicates such as \texttt{connected(L1,
L2)} (i.e., AABBs \texttt{L1} and \texttt{L2} are connected), and the
hierarchical relations in the HLR can be used to construct predicates such as
\texttt{contains(L1, L2)} (i.e. the AABB named \texttt{L1} contains the AABB
named \texttt{L2}).

\subsubsection{Probabilistic Modeling Systems}

Probabilistic models of participants performing tasks in Minecraft (e.g.,
\cite{Pyarelal:2021}) can also make use of the HLR produced by \mcg{} to
construct initial concise internal representations of the task environment.

In general, developing AI agents with machine social intelligence will require
some kind of explicit high-level environment representation to reason about the
beliefs, desires, and intentions of their human partners. \mcg{} treats this
high-level representation as a first-class citizen in its generation framework.

%% file: sections/conclusion.tex
\section{Conclusion}
\label{sec:conclusion}

In this paper, we laid out the motivations for incorporating procedural content
generation into human-machine teaming experiments, and presented our
open-source C++ library, \mcg{}, which integrates low-level content generation
with high-level semantics in order to support human-machine teaming research.
The library provides a set of core components that can be extended and composed
to construct detailed voxel maps while simultaneously generating
machine-readable representations of the environment that can be used by
downstream programs.

\subsection{Limitations}

\subsubsection{Aesthetic concerns}

It is worth noting that generating a structure with non-rectilinear geometry -
for example, something like the Sydney Opera House \cite{sydney_opera_house} -
will be more difficult to do procedurally than manually. In general, if
aesthetic appeal is a significant concern (like it is in the GDMC settlement
generation competition \cite{DBLP:conf/fdg/SalgeGCT18}), an AABB-based
procedural generation approach is likely not an ideal one.  However, in the
context of controlled human-machine teaming experiments, we expect that the
fine-grained control, reproducibility, and scalability afforded by \mcg{} will
outweigh the aesthetic benefits of manual environment generation.

\subsubsection{Programming overhead}

Another potential concern with a procedural generation approach to Minecraft
task environment creation is that it is easier to train people to manually
modify Minecraft environments than to write programs to generate the
environments procedurally. This is not an insignificant concern, especially
when considering that human-machine teaming experiments are often designed in
collaboration with researchers who are not used to writing C++ programs.
However, our stance is that the benefits of using a PCG approach (and the
downsides of manual environment creation) are significant enough to warrant
investing in procedural generation.

\subsection{Future work}

We intend to continue to develop \mcg{} as a part of the ToMCAT project. Along
with making it easier to use and better documented, we will implement
additional AABB-based algorithms to enable researchers to create rich, yet
controlled voxel-based virtual task environments for human-machine teaming
research.

%% file: main.bbl
\begin{thebibliography}{10}
\providecommand{\url}[1]{#1}
\csname url@samestyle\endcsname
\providecommand{\newblock}{\relax}
\providecommand{\bibinfo}[2]{#2}
\providecommand{\BIBentrySTDinterwordspacing}{\spaceskip=0pt\relax}
\providecommand{\BIBentryALTinterwordstretchfactor}{4}
\providecommand{\BIBentryALTinterwordspacing}{\spaceskip=\fontdimen2\font plus
\BIBentryALTinterwordstretchfactor\fontdimen3\font minus
  \fontdimen4\font\relax}
\providecommand{\BIBforeignlanguage}[2]{{%
\expandafter\ifx\csname l@#1\endcsname\relax
\typeout{** WARNING: IEEEtran.bst: No hyphenation pattern has been}%
\typeout{** loaded for the language `#1'. Using the pattern for}%
\typeout{** the default language instead.}%
\else
\language=\csname l@#1\endcsname
\fi
#2}}
\providecommand{\BIBdecl}{\relax}
\BIBdecl

\bibitem{minecraft}
``{Minecraft Official Site},'' \url{https://www.minecraft.net}.

\bibitem{malmo}
\BIBentryALTinterwordspacing
M.~Johnson, K.~Hofmann, T.~Hutton, and D.~Bignell, ``{The Malmo Platform for
  Artificial Intelligence Experimentation},'' in \emph{Proceedings of the
  Twenty-Fifth International Joint Conference on Artificial Intelligence,
  {IJCAI} 2016, New York, NY, USA, 9-15 July 2016}, S.~Kambhampati, Ed.\hskip
  1em plus 0.5em minus 0.4em\relax {IJCAI/AAAI} Press, 2016, pp. 4246--4247.
  [Online]. Available: \url{http://www.ijcai.org/Abstract/16/643}
\BIBentrySTDinterwordspacing

\bibitem{craftassist}
\BIBentryALTinterwordspacing
J.~Gray, K.~Srinet, Y.~Jernite, H.~Yu, Z.~Chen, D.~Guo, S.~Goyal, C.~L.
  Zitnick, and A.~Szlam, ``{CraftAssist: A Framework for Dialogue-enabled
  Interactive Agents},'' \emph{CoRR}, vol. abs/1907.08584, 2019. [Online].
  Available: \url{http://arxiv.org/abs/1907.08584}
\BIBentrySTDinterwordspacing

\bibitem{DBLP:conf/ijcai/GussHTWCVS19}
\BIBentryALTinterwordspacing
W.~H. Guss, B.~Houghton, N.~Topin, P.~Wang, C.~Codel, M.~Veloso, and
  R.~Salakhutdinov, ``{MineRL: A Large-Scale Dataset of Minecraft
  Demonstrations},'' in \emph{Proceedings of the Twenty-Eighth International
  Joint Conference on Artificial Intelligence, {IJCAI} 2019, Macao, China,
  August 10-16, 2019}, S.~Kraus, Ed.\hskip 1em plus 0.5em minus 0.4em\relax
  ijcai.org, 2019, pp. 2442--2448. [Online]. Available:
  \url{https://doi.org/10.24963/ijcai.2019/339}
\BIBentrySTDinterwordspacing

\bibitem{DBLP:journals/corr/abs-2101-11071}
\BIBentryALTinterwordspacing
W.~H. Guss, M.~Y. Castro, S.~Devlin, B.~Houghton, N.~S. Kuno, C.~Loomis,
  S.~Milani, S.~P. Mohanty, K.~Nakata, R.~Salakhutdinov, J.~Schulman,
  S.~Shiroshita, N.~Topin, A.~Ummadisingu, and O.~Vinyals, ``{The MineRL 2020
  Competition on Sample Efficient Reinforcement Learning using Human Priors},''
  \emph{CoRR}, vol. abs/2101.11071, 2021. [Online]. Available:
  \url{https://arxiv.org/abs/2101.11071}
\BIBentrySTDinterwordspacing

\bibitem{DBLP:conf/cig/NguyenRGM17}
\BIBentryALTinterwordspacing
C.~Nguyen, N.~Reifsnyder, S.~Gopalakrishnan, and H.~Mu{\~{n}}oz{-}Avila,
  ``{Automated Learning of Hierarchical Task Networks for Controlling Minecraft
  Agents},'' in \emph{{IEEE} Conference on Computational Intelligence and
  Games, {CIG} 2017, New York, NY, USA, August 22-25, 2017}.\hskip 1em plus
  0.5em minus 0.4em\relax {IEEE}, 2017, pp. 226--231. [Online]. Available:
  \url{https://doi.org/10.1109/CIG.2017.8080440}
\BIBentrySTDinterwordspacing

\bibitem{tomcat}
``{ToMCAT: Theory of Mind-based Cognitive Architecture for Teams},''
  \url{https://ml4ai.github.io/tomcat}.

\bibitem{Bartlett.ea:2015}
\BIBentryALTinterwordspacing
C.~E. Bartlett and N.~J. Cooke, ``{Human-Robot Teaming in Urban Search and
  Rescue},'' \emph{Proceedings of the Human Factors and Ergonomics Society
  Annual Meeting}, vol.~59, no.~1, pp. 250--254, 2015. [Online]. Available:
  \url{https://doi.org/10.1177/1541931215591051}
\BIBentrySTDinterwordspacing

\bibitem{Demir.ea:2018}
\BIBentryALTinterwordspacing
M.~Demir, N.~J. McNeese, and N.~J. Cooke, ``{Dyadic Team Interaction and Shared
  Cognition to Inform Human-Robot Teaming},'' \emph{Proceedings of the Human
  Factors and Ergonomics Society Annual Meeting}, vol.~62, no.~1, pp. 124--124,
  2018. [Online]. Available: \url{https://doi.org/10.1177/1541931218621028}
\BIBentrySTDinterwordspacing

\bibitem{Szlam.ea:2019}
\BIBentryALTinterwordspacing
A.~Szlam, J.~Gray, K.~Srinet, Y.~Jernite, A.~Joulin, G.~Synnaeve, D.~Kiela,
  H.~Yu, Z.~Chen, S.~Goyal, D.~Guo, D.~Rothermel, C.~L. Zitnick, and J.~Weston,
  ``{Why Build an Assistant in Minecraft?}'' \emph{CoRR}, vol. abs/1907.09273,
  2019. [Online]. Available: \url{http://arxiv.org/abs/1907.09273}
\BIBentrySTDinterwordspacing

\bibitem{Huang.ea:2020}
\BIBentryALTinterwordspacing
L.~Huang, J.~Freeman, N.~Cooke, V.~Buchanan, M.~Wood, M.~Freiman,
  J.~Colonna-Romano, and M.~Demir, ``{ASIST Experiment 1 Study
  Preregistration},'' Dec 2020. [Online]. Available: \url{osf.io/zwau9}
\BIBentrySTDinterwordspacing

\bibitem{Corral.ea:2021}
C.~R. Corral, K.~S. Tatapudi, V.~Buchanan, L.~Huang, and N.~J. Cooke,
  ``{Building a Synthetic Environment to Support Artificial Intelligence
  Research},'' in \emph{Proceedings of the 65th Annual Meeting of the Human
  Factors and Ergonomic Society.}, 2021, (under review).

\bibitem{van2013procedural}
\BIBentryALTinterwordspacing
R.~van~der Linden, R.~Lopes, and R.~Bidarra, ``{Procedural Generation of
  Dungeons},'' \emph{{IEEE} Trans. Comput. Intell. {AI} Games}, vol.~6, no.~1,
  pp. 78--89, 2014. [Online]. Available:
  \url{https://doi.org/10.1109/TCIAIG.2013.2290371}
\BIBentrySTDinterwordspacing

\bibitem{git}
\BIBentryALTinterwordspacing
``{git-scm.com}.'' [Online]. Available: \url{https://git-scm.com}
\BIBentrySTDinterwordspacing

\bibitem{Gallistel:1990}
C.~R. Gallistel, \emph{{The organization of learning}}.\hskip 1em plus 0.5em
  minus 0.4em\relax Cambridge, Mass.: Cambridge, Mass. : MIT Press, 1990,
  includes index.

\bibitem{OKeefe.ea:2016}
J.~O'Keefe and L.~Nadel, \emph{{The Hippocampus as a Cognitive Map}}.\hskip 1em
  plus 0.5em minus 0.4em\relax Oxford: Clarendon Press, 2016.

\bibitem{Warren:2019}
W.~H. Warren, ``{Non-Euclidean navigation},'' \emph{Journal of experimental
  biology}, vol. 222, no. Pt Suppl 1, p. jeb187971, 2019.

\bibitem{Peer.ea:2021}
M.~Peer, I.~K. Brunec, N.~S. Newcombe, and R.~A. Epstein, ``{Structuring
  Knowledge with Cognitive Maps and Cognitive Graphs},'' \emph{Trends in
  cognitive sciences}, vol.~25, no.~1, pp. 37--54, 2021.

\bibitem{Maehara.ea:2018}
\BIBentryALTinterwordspacing
H.~Maehara and H.~Martini, ``{Elementary geometry on the integer lattice},''
  \emph{Aequationes mathematicae}, vol.~92, no.~4, pp. 763--800, 2018.
  [Online]. Available: \url{https://doi.org/10.1007/s00010-018-0557-4}
\BIBentrySTDinterwordspacing

\bibitem{DBLP:conf/siggraph/Perlin85}
\BIBentryALTinterwordspacing
K.~Perlin, ``{An image synthesizer},'' in \emph{Proceedings of the 12th Annual
  Conference on Computer Graphics and Interactive Techniques, {SIGGRAPH} 1985,
  San Francisco, California, USA, July 22-26, 1985}, P.~Cole, R.~Heilman, and
  B.~A. Barsky, Eds.\hskip 1em plus 0.5em minus 0.4em\relax {ACM}, 1985, pp.
  287--296. [Online]. Available: \url{https://doi.org/10.1145/325334.325247}
\BIBentrySTDinterwordspacing

\bibitem{santamaria2014procedural}
A.~Santamar{\'\i}a-Ibirika, X.~Cantero, M.~Salazar, J.~Devesa, I.~Santos,
  S.~Huerta, and P.~G. Bringas, ``{Procedural approach to volumetric terrain
  generation},'' \emph{The Visual Computer}, vol.~30, no.~9, pp. 997--1007,
  2014.

\bibitem{green2019organic}
\BIBentryALTinterwordspacing
M.~C. Green, C.~Salge, and J.~Togelius, ``{Organic Building Generation in
  Minecraft},'' \emph{CoRR}, vol. abs/1906.05094, 2019. [Online]. Available:
  \url{http://arxiv.org/abs/1906.05094}
\BIBentrySTDinterwordspacing

\bibitem{lopes2010constrained}
R.~Lopes, T.~Tutenel, R.~M. Smelik, K.~J. De~Kraker, and R.~Bidarra, ``{A
  constrained growth method for procedural floor plan generation},'' in
  \emph{Proc. 11th Int. Conf. Intell. Games Simul}.\hskip 1em plus 0.5em minus
  0.4em\relax Citeseer, 2010, pp. 13--20.

\bibitem{Elliot:2019}
\BIBentryALTinterwordspacing
``{Artificial Social Intelligence for Successful Teams (ASIST)},'' 2019.
  [Online]. Available:
  \url{https://www.darpa.mil/program/artificial-social-intelligence-for-successful-teams}
\BIBentrySTDinterwordspacing

\bibitem{mqtt}
\BIBentryALTinterwordspacing
``{MQTT: The Standard for IoT Messaging}.'' [Online]. Available:
  \url{https://mqtt.org}
\BIBentrySTDinterwordspacing

\bibitem{ihmc}
\BIBentryALTinterwordspacing
``{Florida Institute for Human \& Machine Cognition}.'' [Online]. Available:
  \url{https://www.ihmc.us}
\BIBentrySTDinterwordspacing

\bibitem{pygl_fov}
``{PyGL Field of View},'' \url{https://gitlab.com/cmu_asist/pygl_fov}.

\bibitem{DBLP:conf/acl/JayannavarNH20}
\BIBentryALTinterwordspacing
P.~Jayannavar, A.~Narayan{-}Chen, and J.~Hockenmaier, ``{Learning to execute
  instructions in a Minecraft dialogue},'' in \emph{Proceedings of the 58th
  Annual Meeting of the Association for Computational Linguistics, {ACL} 2020,
  Online, July 5-10, 2020}, D.~Jurafsky, J.~Chai, N.~Schluter, and J.~R.
  Tetreault, Eds.\hskip 1em plus 0.5em minus 0.4em\relax Association for
  Computational Linguistics, 2020, pp. 2589--2602. [Online]. Available:
  \url{https://doi.org/10.18653/v1/2020.acl-main.232}
\BIBentrySTDinterwordspacing

\bibitem{Pyarelal:2021}
\BIBentryALTinterwordspacing
A.~Pyarelal, ``{UArizona ASIST Experiment 1 Preregistration},'' Feb 2021.
  [Online]. Available: \url{osf.io/c926k}
\BIBentrySTDinterwordspacing

\bibitem{sydney_opera_house}
\BIBentryALTinterwordspacing
``{Sydney Opera House: Official Website}.'' [Online]. Available:
  \url{https://www.sydneyoperahouse.com}
\BIBentrySTDinterwordspacing

\bibitem{DBLP:conf/fdg/SalgeGCT18}
\BIBentryALTinterwordspacing
C.~Salge, M.~C. Green, R.~Canaan, and J.~Togelius, ``{Generative design in
  Minecraft {(GDMC):} settlement generation competition},'' in
  \emph{{Proceedings of the 13th International Conference on the Foundations of
  Digital Games, {FDG} 2018, Malm{\"{o}}, Sweden, August 07-10, 2018}},
  S.~Dahlskog, S.~Deterding, J.~M. Font, M.~Khandaker, C.~M. Olsson, S.~Risi,
  and C.~Salge, Eds.\hskip 1em plus 0.5em minus 0.4em\relax {ACM}, 2018, pp.
  49:1--49:10. [Online]. Available:
  \url{https://doi.org/10.1145/3235765.3235814}
\BIBentrySTDinterwordspacing

\end{thebibliography}
